\documentclass{article}
\usepackage{spconf,amsmath,epsfig}

\usepackage{graphics}
\usepackage{graphicx}

\newcommand{\f}{{I\!\!F}}
\newcommand{\g}{{I\!\!E}}

\newcommand{\TRS}{{TRS1027465}}

\title{VOICE BASED SELF HELP SYSTEM: USER EXPERIENCE VS ACCURACY}

\name{Sunil Kumar Kopparapu}
\address{TCS Innovation Lab - Mumbai\\
Tata Consultancy Services, 
Yantra Park, %Subhash Nagar, Pokhran Road 2, 
Thane (West), Maharastra, India.\\
Email: SunilKumar.Kopparapu@TCS.Com}
\begin{document}
%\ninept
%
\maketitle
%\vspace{-0.5in}

%
\begin{abstract}

In general, self help systems are being increasingly deployed by {\em service
based} industries because they are capable of 
delivering better customer service and increasingly the switch is to voice based self help systems because they provide a {\em 
natural} interface for a  human to interact with a machine. 
A speech based self help system ideally needs a speech 
recognition engine to convert spoken speech to text and in addition a language
processing engine to take care of any 
mis-recognitions by the speech recognition engine. Any off-the-shelf speech
recognition engine is generally a combination of acoustic processing
and speech grammar. 
While this is the norm, we believe that ideally a speech recognition application should 
have in addition to a speech recognition engine a {\em separate} language processing engine to give the system better 
performance. In this paper, we discuss ways in which the speech recognition engine and the language 
processing engine can be combined to give a better user experience.

%Any speech based solution requires a speech recognition (SR) engine to convert
%speech to text string and a language processing engine to work on the text
%string. Conventionally language processing is an integral part of the SR
%engine. Oflate Natural language processing engine has been used to enhance the
%quality of output of the SR engine. In this paper we discuss synergistic integration of the language
%processing module of SR engine and a language processing module of Natural
%Language (NL) engine. 
%We emphasize that indeed language processing is required in
%both SR and NL and show that  transferring language processing to either only
%the SR or the NL engine is not optimal.
%We further show how language processing works by building a self help application.

\end{abstract}


\section{Introduction}
\label{sec:intro}

Self help solutions are being increasingly deployed by many service oriented
industries essentially to serve their customer-base any time, any where.
Technology based on artificial intelligence are being used to develop self help
solutions. Typically self help solutions are web based and voice based. Oflate
use of voice based self help solutions are gaining popularity because of the
ease with which they can be used fueled by the significant development in the
area of speech technology.
%Typical deployments are  web agent that provide an interface for
%people to text ƣn their queries and the system responds 
Speech recognition engines are being increasingly used in several applications
with varying degree of success. The reason businesses are 
investing in speech are several. Significant reasons among them are 
the return on investment (RoI) which for speech recognition solutions is 
typically $9-12$ months. In many cases it is as less as $3$
months \cite{www:guide_speech_sol}. In addition, speech solutions 
are  economical and effective
in improving customer satisfaction, operations and workforce 
productivity.
Password resetting using speech 
\cite{www:password}, airline enquiry, talking yellow pages 
\cite{www:tell_me} and more recently in the area of contact centers are some of
the areas where speech solutions have been demonstrated and used.

%Though there has been no {\em eventful }
%change in the way the speech engines function since the day  
%hidden Markov model (HMM) was first used for speech recognition
%there has been a significant improvement in the performance of speech solution. 
The 
increased performance of the speech solution  can be 
primarily attributed to several factors. For example, the work 
in the area of dialog design, language processing have 
contributed to the performance enhancement of the speech solution making them
deployable in addition to the fact that people have become more comfortable
using voice as an interface to transact now. 
The performance of a speech engine is primarily based on 
two aspects, namely,
(a) the acoustic performance and (b) the non-acoustic performance. 
While the change in the 
acoustic performance of the speech engine has increased moderately
 the  mature use of 
non-acoustic aspects have made the speech engine usable in applications; a
combination of this in total  enables good user experience. 

 Any speech based solution requires a spoken
speech signal to be converted into text and this text is further 
processed to derive some form of information from an electronic 
database ({in most practical systems}). The process of converting 
the spoken speech into text is broadly the speech recognition engine 
domain while the later, converting the text into a meaningful text 
string to enable a machine to process it is the domain of natural 
language (NL) processing. In literature, these two have a very thin line 
demarcating them and is usually fuzzy, because the language processing 
is also done in the form of speech grammar in a speech recognition 
engine environment. This has been noted by Pieraccini et al \cite{981383},
where they talk about the complexity involved in integration 
language constraints into a large vocabulary speech recognition
system. They propose to have a limited language capability in the speech
recognizer and transfer the complete language capability to a post processing
unit. Young et al \cite{63344} speak of a 
system 
which combines natural language processing with speech
understanding 
in the context of a problem solving dialog while
Dirk et al \cite{pr_Dirk_05} suggest the use language model to integrate speech
recognition with semantic analysis.
The MIT Voyager speech understanding system 
\cite{ICASSP91-1-178}
interacts with the user through spoken dialog
and the authors describe their attempts at the integration
between the speech recognition and natural language components. 
%They used the
%generation capability of the natural language component to produce a word-pair
%language model to constrain the recognizer's search space, thus improving the
%coverage of the overall system. 
\cite{1273186} talks of combining statistical and knowledge based spoken
language to enhance speech based solution.

  In this paper, we describe a non-dialog based self help system, meaning, a query by the user is  
responded by a single answer by the system; there is no interactive 
session between the machine and the user (as in a dialog based system).
The idea
being that the user queries and the system responds with an answer
assuming that the query is complete in the sense that an answer is {\em
fetchable}.
In the event the query is incomplete the natural language processing (NL) engine
responds with a close answer by making a few assumptions, when required
\cite{DBLP:conf/hci/KopparapuSR07}. 
The NL engine in addition also {\em corrects} any possible speech engine mis-recognition.
In this paper, we make no effort to distinguish 
the differences in the way language is processed in the speech 
recognition module and the natural language processing module. We argue 
that it is optimal ({in the sense of performance of the speech 
solution plus user experience})
to use as a combination of language model in  the speech recognition and natural language 
processing modules.  We first show (Section \ref{sec:background}) that language processing has to be
distributed and can not be limited to either the speech recognition or the
natural language processing engine. We then show how using a readily available 
SR engine and a NL engine \cite{DBLP:conf/hci/KopparapuSR07} how the distribution of
language processing helps in proving better user experience. We describe a
speech based self
self help system in Section \ref{sec:self_help_system} and describe the user experiences in Section \ref{sec:example}. We conclude in Section
\ref{sec:conclusions}.

\section{Background}
\label{sec:background}

For any speech based solution to work in field there are two important
parameters, namely, the accuracy of the speech recognition engine and the
overall user experience. 
While both of these are not 
entirely independent  it is useful to consider them as being independent to be
able to understand the performance of the speech solution and the user
experience associated with the solution.
User experience is measured by the freedom the system gives the user in
terms of (a) who can speak (speaker independent), (b) what can be spoken (large
vocabulary) and (c) how to speak (restricted or free speech) while the speech
recognition accuracies are measured as the ability of the speech engine to
convert the spoken speech into {\em exact} text.

Let $\f$ represent speech recognition engine and 
let  $\g$ be the natural language processing engine. Observe that
  $\f$ converts the
acoustic signal or a time sequence into a string sequence (string of words)
\[
\f : \mbox{time sequence}  \rightarrow \mbox{string sequence} 
\]
while $\g$ processes a string sequence to generate another string sequence.
\[
\g : \mbox{string sequence} \rightarrow \mbox{string sequence}
\]

Let $q_t$ represents the spoken query corresponding to, say, the string 
query  $q_s$ (it can be considered the read version of the written
string of words $q_s$).
Then the operations of the speech and the natural language processing engines
can be represented as
\begin{eqnarray}
 \f(q_t) &=& q_{s'} \;\; \mbox{(speech engine)} \nonumber \\ 
 \g(q_{s'}) &=& q_{s''} \;\; \mbox{(NL processing)}
\end{eqnarray}
Clearly,
the speech recognition engine $\f$ uses  acoustic models ({usually
hidden Markov Model based}) and language grammar which are tightly
coupled to  convert $q_t$ to $q_{s'}$ while the natural language engine
$\g$ operates on $q_{s'}$ and uses {\em only} statistical or knowledge based 
language grammar to convert it into $q_{s''}$. It is clear that the language
processing happens both in $\f$ and $\g$ the only difference being that the
language processing in $\f$ is tightly coupled with the overall functioning of
the speech recognition engine unlike in $\g$.
%\begin{table}
Language processing or grammar used in $\f$ is 
tightly coupled with the acoustic models and hence 
the degree of configurability is very limited  (speech to text). At the same
time language processing
 is necessary to perform {\em reasonable} recognition (speech
recognition performance).
While there is a relatively high
degree of configurability possible in $\g: q_s \rightarrow q_s$  (text to
text).
The idea of any speech based solution is to build $\f$ and $\g$ such that
their combined %\footnote{usually sequential} 
effort, namely, $\g(\f(q_t)) =  q_{s''}$ is such that $q_{s''} \approx q_{s}$.
Do we need language processing in both $\f$ and  $\g$ or is it sufficient to
%\begin{description}
%\item[Option 1:] isolate  $\f$ and $\g$; and have 
(a)  isolate  $\f$ and $\g$; and have 
language processing only in $\g$   or
%\item[Option 2:] 
(b)  
combine all language processing into $\f$ and do away with 
$\g$ completely.
%\end{description}
Probably there is an  optimal combination of $\f$ and $\g$ which produces
a usable speech based solution. 
%It is necessary that both $\f$ and $\g$ need 
%to coexist with {\em some} overlap of language processing
%functionality to produce better (user) interfaces.

An ideal speech recognition system should be able to convert 
$q_t$ into the exact query string $q_s$. Assume that there are three different
types of speech recognition engines. Let the speech recognition engine 
$\f_1$  allow any user to speak anything (speaker independent dictation system);
%\footnote{a typical dictation system};
$\f_2$ be such that it is  $\f_1$ but the performance is 
tuned to a particular person (person dependent) and
 $\f_3$ is such that it is  $\f_2$ additionally constrained in the sense that
it  allows the user to speak from within a restricted grammar. 
Clearly the user
experience is best for $\f_1(x_t) = x^1_{s'}$ (user experience: $\uparrow$)
and worst for $\f_3(x_t) = x^3_{s'}$ (user experience: $\downarrow$) and it
between experience is provided by 
$\f_1(x_t) = x^2_{s'}$ (user experience: $\leftrightarrow$).

Let 
%\[
$
d(x_s,y_s)$ 
be the  
%\stackrel{\Delta}{=}$ 
{distance between the string}
$x_s$ and $y_s$.
%\]
Clearly, $d(x^1_{s'},x_s)$ $>$ $d(x^2_{s'},x_{s})$ $>$ $d(x^3_{s'},x_s)$, the
performance of the speech engine is best for $\f_3$  followed by $\f_2$
followed by $\f_1$. Observe that in terms of user experience it is the reverse.
For the
overall speech system to perform {\em well} the
contribution of $\g$ would vary, namely $\g$ should be able to generate 
$q^1_{s''}$, 
$q^2_{s''}$ and
$q^3_{s''}$ using 
$q^1_{s'}$, 
$q^2_{s'}$ and
$q^3_{s'}$  respectively, so that 
$d(q^1_{s''},q_s)$ $\approx$ $d(q^2_{s''},q_{s})$
$\approx$
$d(q^3_{s''},q_s)$ $\approx$ $0$. The performance of $\g$ has to be 
better to compensate for the {\em poor} performance of $\f$; for example the
performance of $\g_1$ has to be better than the performance of $\g_3$ to
compensate for the poor performance of $\f_1$ compared to $\f_3$.
%What happens if the performance of $\f$ is {\em poor} it is not hard to guess 
%that unless
%$d(q^1_{s'},q_s) < \epsilon$  or $\f(q_t) - q_s < \epsilon$ will
%$\g$ have a role to perform else the input to $\g$ will be too noisy for $\g$
%to perform\footnote{garbage in garbage out situation}.
%Here $\epsilon$ is some measure which determines the minimum performance
%required of the speech recognition engine $\f$. The value of $\epsilon$ depends
%on several issues like the problem at hand and would also depend on the domain
%in which the speech solution is being used etc.

Typically, a $\f_1$ ({ideal user experience}) speech recognition would be categorized by
(a) Open Speech (free speech - speak without constraints),
(b) Speaker independent (different accents, dialects, age, gender) and
(c) Environment independent (office, public telephone). While (a) greatly
depends on the language model used in the speech recognition system, both (b) 
and (c) depend on the acoustic models in the SR.  For $\f_1$ type of system,
the user experience is good but
speech recognition engine accuracies are poor. On the other hand, a typical
$\f_3$ ({bad on user experience}) would be categorized by limiting the
domain of operation 
%(queries related to specific domain insurance agent self help)
and the system would be 
tuned (in other words constrained) to make use of  prior information on expected
type of queries. 
%This is typically achieved by designing language grammar in  SR engine
%by using information available about the typical  expected questions/queries.  

In the next section we describe a voice based self help system which enables us
to tune the language grammar and hence control the performance of the speech
recognition engine.

\section{Voice Based Self Help System}
\label{sec:self_help_system}

Voice based self help system is a speech enabled solution which enables human
users to interact with a machine using their speech to carry out a
transaction. 
To better understand the role of $\f$ and $\g$ in a speech solution we actually
built a voice based self help system. The self help system 
was built using 
the Speech Recognition ($\f$) engine of 
 Microsoft using Microsoft SAPI SDK \cite{ms_sapi} and the 
 Language Processing ($\g$) module was developed in-house
\cite{DBLP:conf/hci/KopparapuSR07}. 

In general, insurance agents act as  intermediaries between  
the insurance company (service
providing company) and their clients (actual insurance seekers).
Usually, the insurance agents keep track of information of their clients
(policy status, maturity status, change of address request among other things)
by being in touch with the insurance company.
In the
absence of a self help system, the insurance agents got information by
speaking to live agents at a call center run by the insurance company.
The reason for
building a self help system was to enable the
insurance company to lower the use of call center usage and additionally 
providing dynamic
information needed by agents; both this together 
provide better customer service. 
The automated self help system, enabled answering queries of an insurance
agent.
%associated with a certain insurance company. 
\begin{figure}[h]
\centerline{\includegraphics[width=.45\textwidth]{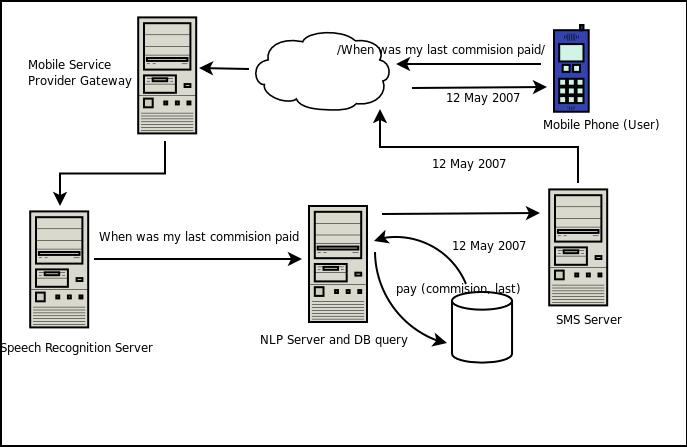}}
\caption{Block Diagram of a Self Help System}
\label{fig:shs_block}
\end{figure}
Figure \ref{fig:shs_block} shows a high level 
functional block
representation of the self help system.
The user
(represented as a mobile phone in Figure \ref{fig:shs_block}) calls a
predetermined number 
and speaks his queries to get information. The speech recognition engine
converts the spoken query (speech signal) into text; this text is operated upon by the
natural language processing block. This processed string is then used to fetch an answer
from the database. The response to this query, which is a text string, 
is (a) spoken out to the user
using a text to speech engine and
(b) alternatively is sent to the user as a SMS. 
We used Kannel, an open source WAP and SMS gateway to send the answer string 
as SMS \cite{web_kannel}.

The self help  system was designed to cater to
\begin{enumerate}
\item different kinds of information sought by insurance agent
\begin{enumerate}
\item on behalf of their clients (example, {\em What
is the maturity value of the policy \TRS}) and 
\item themselves (example, {\em When was my last commission paid?}).
\end{enumerate}
%\item  different 
%\begin{enumerate}
\item  different accents,
\item  handle different complexity of queries and
%\end{enumerate}
\item Additionally the system should be able to accept 
natural English query. 
\item   different ways in which same queries can be asked,
%Namely, 
%the system should allow querying in several different ways (natural
%language). 
Examples:
\begin{enumerate}
\item  Surrender value of policy \TRS?
\item  What is the surrender value of policy \TRS?
\item  Can you tell me surrender value of policy \TRS?
\item  Please let me know the surrender value of policy \TRS?
\item  Please tell me surrender value of policy \TRS?
\item  Tell me surrender value of policy \TRS?
\item  My policy is \TRS. What is its surrender value?
\end{enumerate}
should all be understood as being queried for the 
\begin{quote}
{{\em ... surrender\_value} of the {\em policy \TRS ....}}
\end{quote}
\end{enumerate}

Note that the performance of
the speech recognition engine is controlled by the speech 
grammar (see Figures \ref{fig:no_grammar}, \ref{fig:liberal_grammar},
\ref{fig:constrained_grammar} for examples of speech grammar) that driveѕ the
speech recognition engine.  The speech grammar is used by the speech engine
before converting the spoken acoustic signal into a text string. 
In Section \ref{sec:example} we show
how the performance of the speech engine can be 
controlled by varying the speech grammar.

\section{Experimental Results}
\label{sec:example}

We built three versions of the self help system with varying degrees of processing 
distributed in $\f$ and $\g$, namely, 
\begin{enumerate}
\item  $\f_1$
 has {no} grammar
 (Figure \ref{fig:no_grammar}), giving a very high degree of freedom to the
user as to what they can ask, giving them scope to ask invalid queries.
\item $\f_2$
 (Figure \ref{fig:liberal_grammar})
 has liberal grammar; more processing in $\g$
and 
\item $\f_3$
 has a constrained grammar
 (see Figure \ref{fig:constrained_grammar}) which constraints the 
flexibility of what the user can say 
\end{enumerate}
For example, $\f_1$ grammar would validate even an out of domain query
like {\em What does this system do?} in one dimension and an incorrect query
like {\em What is last paid commission address change?}. On the other extreme a
$\f_3$ grammar would only recognize queries like {\em What is the surrender
value of Policy number} or {\em Can you please tell me the maturity value of
Policy number} and so on.
Note that the constrained grammar $\f_3$ gives 
a very accurate speech recognition because the speaker speaks what the speech
engine expects this in turn puts very less load in terms of processing on
$\g$.  

For the experimental setup, the $\f_1$ grammar
 generated a total of $27$ possible
queries of which only $3$ were not responded by the $\f_1$, $\g$ combined
system. On the other hand for a grammar of type $\f_3$
a total of $357$ different queries that the user could ask possible (very
high degree of flexibility to the user). Of these only a total of $212$ queries were
valid in the sense that they were meaningful and could be answered by the
$\f_3$, $\g$ system the rest, $145$, were processed by $\g$ but were not
meaningful and hence an answer was not provided. 
The performance of $\f_2$ grammar was in between these two cases
producing a total of $76$ possible queries that the user could ask, of which 
$20$ were invalided by the $\f_2$, $\g$ combine. 

\begin{figure}[h]
%\begin{small}
\begin{verbatim}
<GRAMMAR>
 <RULE NAME="F_1" TOPLEVEL="ACTIVE">
  <RULEREF NAME="DonotCare"/>
 </RULE>
</GRAMMAR>
\end{verbatim}
%\end{small}
\caption{$\f_1$: No grammar; the speaker can speak anything.}
\label{fig:no_grammar}
\end{figure}

\begin{figure}[h]
%\begin{small}
%\fbox{ \hbox{
\begin{verbatim}
<GRAMMAR>
 <RULE NAME="F_2" TOPLEVEL="ACTIVE">
  <RULEREF NAME="DonotCare"/>
  <RULEREF NAME="KeyConcept"/>
  <RULEREF NAME="DonotCare"/>
  <RULEREF NAME="KeyWord"/>
  <RULEREF NAME="DonotCare"/>
 </RULE>
 <RULE NAME="KeyConcept">
  <P> Surrender Value </P>
  <P> Maturity Value </P>
  <P> ... </P>
  <P> Address Change </P>
 </RULE>
 <RULE NAME="KeyWord">
  <P> Policy Number </P>
  <P> ... </P>
  <P> ... </P>
 </RULE>
</GRAMMAR>
\end{verbatim}
%}}
%\end{small}
\caption{$\f_2$: Liberal grammar: Some restriction on the user.}
\label{fig:liberal_grammar}
\end{figure}

%{$\f_1$ - constrained grammar}

\begin{figure}[h]
%\begin{small}
\begin{verbatim}
<GRAMMAR>
 <RULE NAME="F_3" TOPLEVEL="ACTIVE">
  <o> <RULEREF NAME="StartTag"/> </o>
  <RULEREF NAME="KeyConcept"/>
  <o> of <o> the </o> </o>
  <o> in <o> the </o> </o>
  <RULEREF NAME="KeyWord"/>
  <o> <RULEREF NAME="EndTag"/> </o>
 </RULE>
 <RULE NAME="StartTag">
  <P> What is the </P>
  <P> Please send me </P>
  <P> Can you please send me</P>
  <P> Can you tell me </P>
 </RULE>
 <RULE NAME="KeyConcept">
  <P> Surrender Value </P>
  <P> Maturity Value </P>
  <P> ... </P>
  <P> Address Change </P>
 </RULE>
 <RULE NAME="KeyWord">
  <P> Policy Number </P>
  <P> ... </P>
  <P> ... </P>
 </RULE>
 <RULE NAME="EndTag">
  <P> Thank You </P>
  <P> ... </P>
 </RULE>
</GRAMMAR>
\end{verbatim}
%\end{small}
\caption{$\f_3$: Constrained grammar - speaker is highly constrained in what he
can speak.}
\label{fig:constrained_grammar}
\end{figure}

\section{Conclusions}
\label{sec:conclusions}

The performance of a voice based self help solution has two components; user
experience and the performance of the speech engine in converting the spoken
speech into text. It was shown that  $\f$ and $\g$ can be used jointly to come up
with types of self help solutions which have varying effect on the user
experience and performance of the speech engine. Further, we showed that on one
hand by controlling
the language grammar one could provide better user experience but the performance
of the speech recognition became poor while on the other hand when the
grammar was such that the performance of speech engine was good the 
user experience became poor. This shows that there is a balance between the speech
recognition accuracy and user experience that is to be maintained by people
who design voiced based self help systems so that both the speech recognition
accuracy is good without sacrificing the user experience. 
%We built a working system which has the following
%advantages, (a)
%$24 \times 7$ (support for frequent asked queries)
%(b) Automated  self service (reducing call volume)
%(c) Accuracy and consistency in call experience
%(d) Automated speech and text (SMS) response (provides a choice)
%(e) Ability to handle multiple levels of complexity

\bibliographystyle{IEEEbib}
\bibliography{nl_sr_integration_icon08}

\begin{thebibliography}{10}

\bibitem{www:guide_speech_sol}
Daniel Hong,
\newblock ``An introductory guide to speech recognition solutions,'' Industry
  white paper by Datamonitor, 2006.

\bibitem{www:password}
Microsoft Research,
\newblock ``Microsoft speech - solutions: Password reset,''
  http://www.microsoft.com/ speech/ solutions/ pword/ default.mspx, 2007.

\bibitem{www:tell_me}
Tellme,
\newblock ``Every day info,'' http://www.tellme.com/ products/ TellmeByVoice,
  2007.

\bibitem{981383}
Roberto Pieraccini and Chin-Hui Lee,
\newblock ``Factorization of language constraints in speech recognition,''
\newblock in {\em Proceedings of the 29th annual meeting on Association for
  Computational Linguistics}, Morristown, NJ, USA, 1991, pp. 299--306,
  Association for Computational Linguistics.

\bibitem{63344}
S.~L. Young, A.~G. Hauptmann, W.~H. Ward, E.~T. Smith, and P.~Werner,
\newblock ``High level knowledge sources in usable speech recognition
  systems,''
\newblock {\em Commun. ACM}, vol. 32, no. 2, pp. 183--194, 1989.

\bibitem{pr_Dirk_05}
Dirk Buhler, Wolfgang Minker, and Artha Elciyanti,
\newblock ``Using language modelling to integrate speech recognition with a
  flat semantic analysis,''
\newblock in {\em 6th SIGdial Workshop on Discourse and Dialogue}, Lisbon,
  Portugal, September 2005.

\bibitem{ICASSP91-1-178}
Victor~W. Zue, James Glass, David Goodine, Hong Leung, Michael Phillips, Joseph
  Polifroni, and Stephanie Seneff,
\newblock ``{Integration of Speech Recognition and Natural Language Processing
  in the MIT Voyager System},''
\newblock in {\em Proc. ICASSP}, 1991, vol.~1, pp. 713--716.

\bibitem{1273186}
Ye-Yi Wang, Alex Acero, Milind Mahajan, and John Lee,
\newblock ``Combining statistical and knowledge-based spoken language
  understanding in conditional models,''
\newblock in {\em Proceedings of the COLING/ACL on Main conference poster
  sessions}, Morristown, NJ, USA, 2006, pp. 882--889, Association for
  Computational Linguistics.

\bibitem{DBLP:conf/hci/KopparapuSR07}
Sunil~Kumar Kopparapu, Akhlesh Srivastava, and P.~V.~S. Rao,
\newblock ``Minimal parsing key concept based question answering system,''
\newblock in {\em HCI (3)}, Julie~A. Jacko, Ed. 2007, vol. 4552 of {\em Lecture
  Notes in Computer Science}, pp. 104--113, Springer.

\bibitem{ms_sapi}
Microsoft,
\newblock ``Microsoft {S}peech {API},''
\newblock {\em http://msdn.microsoft.com/en-us/library/ms723627(VS.85).aspx},
  Accessed Nov 2008.

\bibitem{web_kannel}
Open Source,
\newblock ``Kannel: Open source {WAP} and {SMS} gateway,''
\newblock {\em http://www.kannel.org/}, Accessed Nov 2008.

\end{thebibliography}

\end{document}